\documentclass[accepted]{uai2026}

\usepackage[american]{babel}

\usepackage{natbib}
\usepackage{mathtools}
\usepackage{siunitx}
\usepackage{booktabs}
\usepackage{tikz}

\usepackage{amsthm}
\usepackage{amsmath}
\usepackage{amsfonts}
\usepackage{bm}
\usepackage{cleveref}
\usepackage{color}
\usepackage[inline]{enumitem}
\usepackage{float}
\usepackage{multirow}
\usepackage{subcaption}
\usepackage{tabularx} 
\usepackage{stfloats}

\title{Bayesian Lottery Ticket Hypothesis}

\author[1]{\href{mailto:<nicholas.kuhn@kit.edu>?Subject=Bayesian Lottery Ticket Hypothesis}{Nicholas Kuhn}{}}
\author[1]{Arvid Weyrauch}
\author[1]{Lars Heyen}
\author[1]{Achim Streit}
\author[1,2]{Markus Götz}
\author[1]{Charlotte Debus}
\affil[1]{%
    Scientific Computing Center (SCC)\\
    Karlsruhe Institute of Technology (KIT)\\
    Karlsruhe, Germany
}
\affil[2]{%
    Helmholtz AI\\
    Eggenstein-Leopoldshafen, Germany
}
  
\begin{document}
\maketitle

\begin{abstract}
    Bayesian neural networks (BNNs) are a useful tool for uncertainty quantification, but require substantially more computational resources than conventional neural networks.
    For non-Bayesian networks, the Lottery Ticket Hypothesis (LTH) posits the existence of sparse subnetworks that can train to the same or even surpassing accuracy as the original dense network.
    Such sparse networks can lower the demand for computational resources at inference, and during training. The existence of the LTH and corresponding sparse subnetworks in BNNs could motivate the development of sparse training algorithms and provide valuable insights into the underlying training process.
    Towards this end, we translate the LTH experiments to a Bayesian setting using common computer vision models. We investigate the defining characteristics of Bayesian lottery tickets, and extend our study towards a transplantation method connecting BNNs with deterministic Lottery Tickets.
    We generally find that the LTH holds in BNNs, and winning tickets of matching and surpassing accuracy are present independent of model size, with degradation at very high sparsities. However, the pruning strategy should rely primarily on magnitude, secondly on standard deviation. Furthermore, our results demonstrate that models rely on mask structure and weight initialization to varying degrees.
\end{abstract}

\section{Motivation}

Bayesian neural networks (BNN) are a powerful, yet complex tool for uncertainty quantification (UQ) in neural networks. Conventional neural network (NN) predictions lack confidence estimations that directly map to the prediction uncertainty, e.g., when outputting class prediction probabilities. To overcome this Bayesian principles can be leveraged to promote the model weights from fixed values to distributions, thereby facilitating model deployment in real-world safety-critical applications~\citep{aidriving}. 
One approach towards BNNs is mean-field variational inference (VI)~\citep{saul1996meanfieldtheorysigmoid,blundell2015}, which approximates the true distribution of the model weights with a parameterized variational distribution.
During model training, the parameters of this distribution are optimized such that the output distribution of the model matches the distribution of the training data. In addition to UQ, VI has other interesting features: it can increase a model's generalization capability and decrease overfitting, making BNNs a better choice for small datasets~\citep{izmailov_what_2021}. 

These advantages come with the downside of an increased computational demand:
For one, describing the weights with a parameterized distribution increases the number of free model parameters substantially, thus augmenting model size and corresponding memory demand. 
Further, the distributional nature of the model formulation increases the number of computational operations, i.e., FLOPs. The cost for forward and backward passes due to evaluating multiple samples multiplies the training and evaluation cost beyond that of a deterministic NN. Both the per-iteration computations and the number of samples needed for stable predictions grow with the number of parameters and with the complexity of the posterior. Consequently, large-scale Bayesian models are very challenging to train on consumer hardware~\citep{practicalVariational,stochasticVariational}.

In conventional non-Bayesian NN sparsity has emerged as an effective approach to reducing computational load and memory footprint~\citep{hoefler_sparsity_2021}, given the fact that most parameters of a trained model are close to zero.
From a technical point of view, sparsity in DNNs means that the weight matrices representing a layer carry a large number of zeros, i.e. they are sparse matrices. Only non-zero parameters contribute to a layers output, and zero parameters can therefore be removed from the matrix. This is called pruning, and is usually done after training. A retraining recovers predictive performance lost during pruning.
Pruning BNNs is generally feasible to reduce the computational demand for inference deployment, and has been explored in previous works~\citep{blundell2015}. However, it does not alleviate the far greater problem of computational load for BNN model training, for which a priori knowledge is required on which weights can be removed from the network.

Dynamic pruning methods for training conventional NNs in a sparse manner have been proposed, all of which are founded in the observation of the Lottery Ticket Hypothesis (LTH)~\citep{lotteryticket}.
LTH postulates the existence of subnetworks that train to the same accuracy (matching) as the full network. These so-called Lottery Tickets (LTs) combine the efficiency of a sparse network with the accuracy of a dense one. They are found through Iterative Magnitude Pruning (IMP), in which models undergo a train-prune-reset cycle to find the pair of initial weights and pruning mask.

To evaluate whether such algorithms for training sparse BNNs are viable, this study aims to investigate whether the LTH also holds in BNNs. This question has already garnered interest in the communtiy~\cite{noauthor_bayes_2025}.
If a performant sparse subnetwork also exists in a BNN, both the number of model parameters and computational operations could be reduced, making each training and evaluation step computationally cheaper. Performing posterior inference in the forward pass in a lower-dimensional parameter space can also improve mixing and convergence of Markov Chain Monte Carlo (MCMC)~\citep{neal_mcmc_2011} and variational methods~\citep{izmailov_what_2021}. As a result, one can achieve an equal prediction quality with fewer weight samples, and each sample is less expensive to obtain, enabling the practical training of sparse, large-scale BNNs. 

We start by translating the original LTH experiment into a Bayesian setting. Na\"ively, there is no reason to assume that LTs cannot be found in BNNs. However, the concrete manifestation may differ from the deterministic case. The role of initialization may be less straightforward: since Bayesian weights are drawn from distributions rather than fixed values, the stability and reproducibility may be influenced by the choice of prior or the variance of the variational posterior. Architectural differences may additionally play a role, with convolutional BNNs possibly exhibiting more structured sparsity on the layer-level, while attention-based BNNs may show pruning patterns related to their attention-Multilayer Perceptron (MLP) stack architecture.

To uncover these potential differences, we study LTH in BNNs by applying IMP with different pruning strategies in a set of Bayesian-turned computer vision networks, including both convolutional and attention-based models, and compare them to their non-Bayesian counterparts.
Further, we investigate the characteristics that define a Bayesian LT, by analyzing the sparsity patterns on a global and per-layer level. 
Motivated by the computation cost associated with obtaining Bayesian LTs, we investigate a transplantational strategy in which non-Bayesian LTs are transferred into BNNs. Since training via VI introduces an additional computational burden, delaying VI-based optimization to later stages of the iterative pruning offers the potential to reduce overall compute requirements. We therefore examine whether such transplantation can preserve the performance advantages observed for Bayesian LTs compared to randomly pruned BNNs, while improving computational efficiency and maintaining the calibration benefits of BNNs.
We provide ablational studies on model size, learning rate schedule and learning rate rewinding~\cite{renda_comparing_2020}.

\section{Related Work}

\paragraph{Sparsity} 
Sparsity arises naturally in overparametrized networks, which are often trained with far more weights than are strictly necessary to represent the target function~\citep{han_deep_2016}. By removing this redundancy, sparse models can achieve benefits such as reduced memory footprint, faster inference, and potentially improved generalization. A wide variety of techniques have been developed to induce sparsity in NNs, like explicit regularization~\citep{shen_sparse_2024} to encourage weights to shrink toward zero, or low-rank factorization~\citep{sui_elrt_2024} and sparse convolutions~\citep{elsen_fast_2019}.
\textbf{Pruning} has emerged as one of the most widely studied and practically successful strategies~\citep{hoefler_sparsity_2021}. Pruning techiques start from a dense, trained network and progressively remove parameters deemed unimportant according to some criterion.
\textbf{The Lottery Ticket Hypothesis} marks a special case of pruning by demonstrating the existence of well-performing sparse networks, without the need for a dense training phase commonly used in pruning schemes. Through pruning, a mask for the specific initialization of the weights is found, the mask and initialization form a subnetwork which can match the performance of the dense network.
The LTH has been widely tested on various architectures and task setups~\citep{liu_survey_2024}. Most notably, winning tickets have been demonstrated to exist in linear~\citep{lotteryticket}, convolutional~\citep{lotteryticket}, LSTM~\citep{yu_playing_2020}, attention~\citep{chen_earlybert_2021}, and graph~\citep{chen_unified_2021} layers, as well as in supervised, unsupervised, and reinforcement learning problems~\citep{lotteryticket,itahara_lottery_2020,yu_playing_2020}. Additionally, research has extended to create even better performing LTs~\citep{renda_comparing_2020} in bigger models~\citep{frankle_stabilizing_2020}, or with less compute needed~\citep{wang_picking_2020,lee_snip_2019} by pruning before training.

\paragraph{Bayesian Neural Networks} BNNs add an additional layer of complexity to the application of deep learning, by promoting the model weights from fixed values to distributions.
A common approach to train BNNs is \textbf{Variational Inference (VI)}, which approximates the true distribution of the model weights with a parameterized variational distribution~\citep{practicalVariational,stochasticVariational}. Early VI for neural networks used a fully factorized Gaussian posterior, because of its low memory cost. Mean-field VI, with reparametrization-based approaches~\citep{blundell2015,kingma_variational_2015}, make training large BNNs tractable, but are known to underestimate posterior variance and struggle with multimodality induced by parameter symmetries. A parallel line of work uses stochastic-gradient MCMC~\citep{nemeth_stochastic_2019,wu_functional_2024,kim_learning_2024} and Laplace and second-order approximations~\citep{faller_optimal_2025}, which trade the computational cost for better posterior fidelity. 

The prior is a major lever for inducing sparsity in BNNs. The spike-and-slab prior, a hierarchical two-component mixture prior, explicitly encodes that some weights should be exactly zero~\citep{jantre_spike-and-slab_2025}, alternative sparsity-promoting priors include the horseshoe prior, log-uniform or heavy-tailed priors~\citep{ghosh_model_2017,carvalho_handling_2009,li_training_2024}, which encourage many small weights and few large ones. Work on Bayesian compression combines these priors with structured pruning schedules to achieve highly sparse predictive models~\citep{cherief-abdellatif_convergence_2019}. Other, more general pruning work has focused on posterior pruning~\citep{blundell2015,mathew_pruning_2023}, and uncertainty-aware pruning criteria~\citep{ko_magnitude_2019}.


\section{Background}
\paragraph{Bayesian Neural Networks}

The Bayesian approach to neural networks addresses the lack of epistemic uncertainty representation by placing a prior distribution over the network weights, thereby treating the weights as probabilistic distributions rather than deterministic parameters. A common choice is to model each weight \( w_i \) as a Gaussian distribution:
\[
w_i \sim \mathcal{N}(\mu_i, \sigma_i^2).
\]
However, computing the posterior distribution \( p(\mathbf{w} \mid \mathcal{D}) \) over the weights given data \( \mathcal{D} \) is generally intractable due to the high-dimensional and nonlinear nature of neural networks. \textbf{Variational inference} offers a tractable approximation by positing a simpler variational distribution \( q(\mathbf{w} \mid \theta) \), parameterized by \( \theta \), and minimizing the Kullback-Leibler (KL) divergence between the variational posterior and the true posterior.

This is typically done by maximizing the evidence lower bound (ELBO), given by:
\[
\mathcal{L}(\theta) = \mathbb{E}_{q(\mathbf{w} \mid \theta)} \left[ \log p(\mathcal{D} \mid \mathbf{w}) \right] - \mathrm{KL}(q(\mathbf{w} \mid \theta) \,\|\, p(\mathbf{w})).
\]
The first term encourages the variational weights to explain the observed data, while the second term regularizes the variational distribution to stay close to the prior. The quality of prediction uncertainty can be assessed through the Mean Absolute Calibration Error (MACE), given by:

\[
\mathrm{MACE} = \frac{1}{B} \sum_{b=1}^B \frac{1}{|S_b|} \left| \sum_{i \in S_b} \left( \mathbb{I}\!\left(\hat{y}_i = y_i\right)  - \hat{p}_i\right) \right|
\]
which is a measure for the discrepancy between predicted confidence and empirical accuracy.

Advances in variational inference, such as Bayes by Backprop~\citep{blundell2015} and local reparameterization tricks~\citep{kingma_variational_2015}, have made it feasible to apply Bayesian methods to large-scale models like ResNet~\citep{he2015deepresiduallearningimage}, enabling uncertainty-aware deep learning at scale.


\paragraph{Iterative Magnitude Pruning} IMP as proposed in the Lottery Ticket Hypothesis, is an algorithm where a network is first trained for a number of epochs to obtain a set of learned weights \( \mathbf{w} \). Then, weights are given a score, defined by \( s_{m} = |w| \), and the lowest-scoring weights are pruned. Pruning is implemented by constructing a binary mask \( \mathbf{m} \in \{0,1\}^d \), where \( d \) is the number of parameters, such that:
\[
\mathbf{w}_{\text{pruned}} = \mathbf{m} \odot \mathbf{w},
\]
with \( \odot \) denoting element-wise product. The non-zero entries of \( \mathbf{m} \) correspond to the ``surviving'' weights. 

Importantly, the remaining weights are then reset to their original initialization values, denoted \( \mathbf{w}_0 \). The resulting winning ticket is defined by the pair \( \mathbf{w}_{\text{ticket}, N} = (\mathbf{m_N}, \mathbf{w}_0) \); the train-prune cycle is repeated \( N \) times, achieving higher sparsities at each level, building up a very sparse winning ticket.
Empirical results show that both components, sparsity pattern \( \mathbf{m} \) and initialization \( \mathbf{w}_0 \), are crucial~\citep{lotteryticket}.
Further, some advancements on the study of the LTH have been made, e.g., LTs have been found in other architectures and settings~\citep{chen_unified_2021,chen_earlybert_2021}, but the time-consuming iterative process is still necessary.

\section{Experimental Setup}
The objective of our experiments is to translate the IMP experiment of \citet{lotteryticket} to a Bayesian setting. For that, we implement and train BNNs using mean-field VI, apply pruning in various forms, reset and re-train them to achieve the same training structure as in the LTH. We then compare the resulting predictive performance to the classical non-Bayesian counterparts that underwent IMP following the same procedure. 

\paragraph{Dataset and Models} 
For our evaluation, we chose the classical computer vision models 
ResNet-18~\citep{he2015deepresiduallearningimage}, VGG~\citep{simonyan2015deepconvolutionalnetworkslargescale}, and VisionTransformer~\citep{dosovitskiy_image_2021}, which are all trained on the task of image classification on the CIFAR10~\citep{krizhevsky2009learning} dataset, which comprises 50k training and 10k testing images from ten different classes.
Next to the standard non-Bayesian versions of these models, which serve as a baseline, we implement and train Bayesian variants using mean-field VI.
The common random horizontal flip and cropping data augmentations are applied to the training images~\citep{lotteryticket}. This makes the model and dataset choice equal to the original LTH experiment, with the addition of the VisionTransformer model.

\paragraph{Implementation of Bayesian models}
Every linear and convolutional layer in the models is replaced by its Bayesian counterpart.
Means are initialized according to the Kaiming uniform initialization, and standard deviations are initialized to 1\% of the Kaiming scale.
We forward each batch ten times with differently sampled weights.
The final predictions are averaged over the output samples.
We apply a temperature scaling with $T=0.1$ to the KL loss to inhibit posterior collapse~\citep{wenzel_how_2020}. 

\paragraph{Training Curriculum}
For every run, non-Bayesian and Bayesian, we use the following shared training parameters. Everything non-Bayesian-related and not explicitly stated is kept constant, i.e. only additional parameters introduced in a Bayesian variant are changed. The models are trained for 160 epochs, using the ADAM optimizer with a learning rate of \( 0.001 \) and a learning rate schedule to drop the learning rate by \( 0.1 \) at epochs \( 80 \) and \( 120 \). The weight decay is set to \( 0.0001 \). The maximum achieved test accuracy and MACE is recorded. After training, all convolutional and linear layers except the final classification layer are globally pruned with a pruning rate of \( 0.2 \). Resetting the weights to their original initialization, we repeat this training and prune for 20 levels of sparsity. In the final level, only \( \sim 1.2\% \) of the total parameters are non-zero.

\paragraph{Hardware and Software}
\label{hard_software}
All experiments were run on a compute node equipped with 12 Intel Xeon Platinum 8368 CPUs and four NVIDIA A100 or H100 GPUs as accelerators with 40GB and 96GB VRAM, respectively, connected via NVLink.
All experiments used Python 3.11.7 with \texttt{CUDA}-enabled \texttt{PyTorch} 2.4.0~\citep{paszke2018pytorch}. We use \href{https://github.com/RAI-SCC/torch_blue}{torch-blue}~\cite{weyrauch_torch_blue_2026} for the implementation of Bayesian models.
The source code for our experiments is publicly available\footnote{\footnotesize{\url{https://anonymous.4open.science/r/open_lth-4E2F/}}}.

\section{Pruning in BNNs}
\label{sec:pruning-bnn}
\begin{figure*}[!t]
    \centering
    \includegraphics[width=0.33\linewidth]{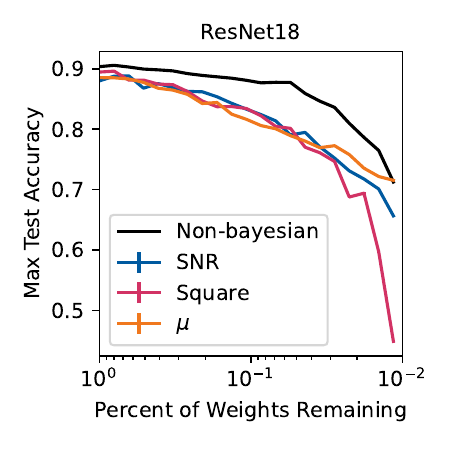}%
    \includegraphics[width=0.33\linewidth]{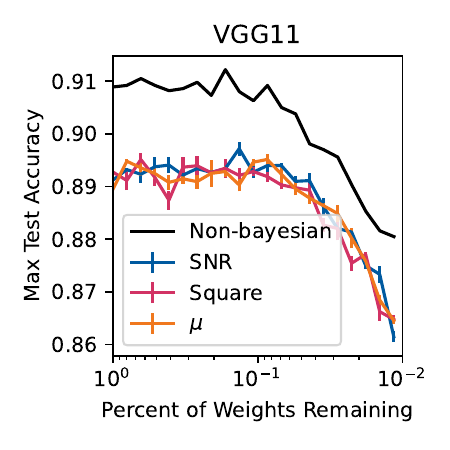}%
    \includegraphics[width=0.33\linewidth]{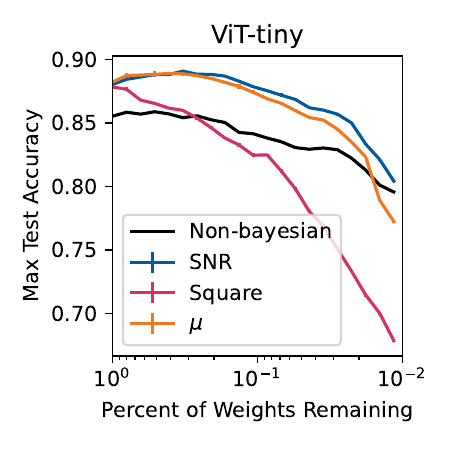}
    \includegraphics[width=0.33\linewidth]{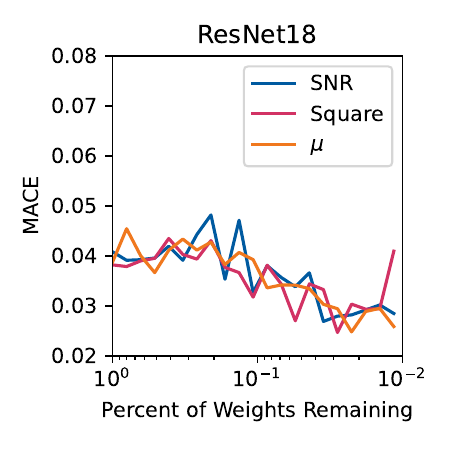}%
    \includegraphics[width=0.33\linewidth]{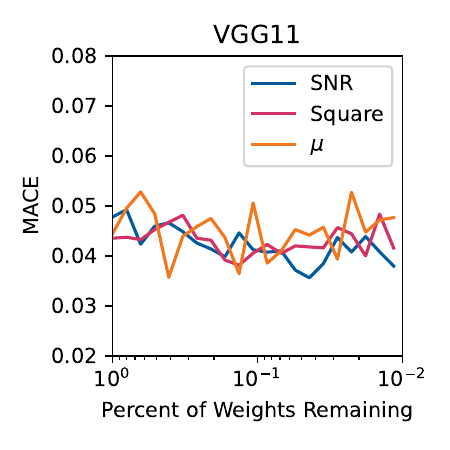}%
    \includegraphics[width=0.33\linewidth]{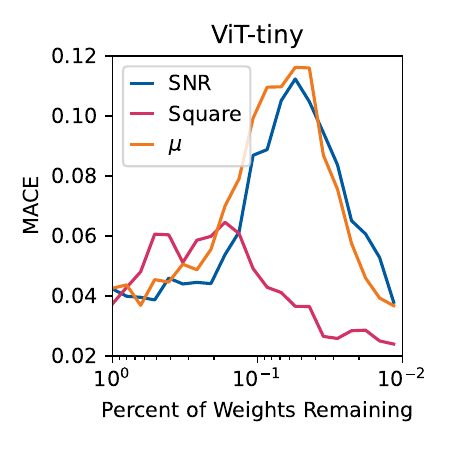}
    \caption{Test accuracy and MACE vs. percentage of weights remaining for ResNet18, VGG11 and ViT-tiny models trained on CIFAR10. Shown in color are the different scoring functions for pruning; in black are the non-Bayesian models pruned by magnitude. The x-axis is shown on a logarithmic scale; the y-axis is adjusted to include the full range of observed accuracies.}
    \label{fig:pruning-comparison}

\end{figure*}

Pruning in BNNs differs in the scoring function used from pruning in deterministic networks. In the original experiments for the LTH, the pruning function $s_m $, defined above, was used. In BNNs, pruning decisions can account for both the mean \( \mu \) and the uncertainty \( \sigma \) associated with each weight. When pruned, a weight has \( (\mu, \sigma) = (0, 0)\).

For our first experiment, we apply three different kinds of unstructured pruning approaches to the Bayesian models. The \textit{signal-to-noise} score assigns a value of
\(
s_{SNR } = \frac{|\mu_|}{\sigma}
\)
to every weight \( w \) with parameters \( (\mu, \sigma) \), and preferably prunes weights that are close to zero and ``noisy'', e.g., have a high standard deviation~\citep{blundell2015}.
The \textit{squared-sum} (Square) score is defined by
\(
s_{ss} = \sqrt{\mu^2 + \sigma^2}.
\)
Similar to \( s_{SNR} \), weights with low \( \mu \) are pruned, but instead, weights with a low standard deviation are preferred, intuitively corresponding to weights where the network is ``sure'' they should be close to zero.
The \textit{mu-magnitude} ($ \mu $) score, defined by
\(
s_{mm} = |\mu|
\)
disregards the standard deviation. It is motivated by the fact that there is no intuition a priori about the value of \( \sigma \) besides the prior matching, and pruning should therefore be independent of its value.
The baseline non-Bayesian model is achieved by pruning based on $s_m$, keeping all other variables equal. 

Figure~\ref{fig:pruning-comparison} shows the recorded accuracy over the percentage of retained weights. Pruning ResNet18, we find a similar behaviour in the baseline as for \(s_{SNR} \text{ and } s_{mm}\), with a gradual accuracy degradation over increasing sparsity. In the high sparsity regime~\({(>98\%)} \), performance degrades quickly after pruning with \(s_{ss} \) and in the baseline.
Pruning VGG11, we find again a similar behaviour in the Bayesian versus the non-Bayesian model, with a decline in performance in the high-sparsity regime, but below 90\% sparsity, all models perform equally well as their dense counterparts. There is no discernible difference between the different pruning scoring functions.
Pruning ViT-tiny, i.e. the attention-based model, the Bayesian variants outperform the non-bayesian model, and $s_{SNR}$ and $s_{\mu}$ exhibit an increase in performance pruning to 50\% sparsity, and only then a decline above 90\% sparsity. The $s_{ss}$ scoring function produces unsatisfactory results with a stark constant decline when pruning.
We therefore do not consider $s_{ss}$ an accurate replacement for magnitude pruning. The value of the standard deviation of a weight should be taken into account when pruning, but the magnitude of the mean can already provide a good performance.
Even though only a few sparsity levels greater than zero achieve surpassing accuracy, the similarity across all levels as compared to the baseline, seems to imply that the LTH also holds in a Bayesian setting. We will refer to $ s_{SNR} $ as IMP when referring to Bayesian models.

Error bars are included in the test-set accuracy plots of the Bayesian models in Figure~\ref{fig:pruning-comparison}, obtained from repeated stochastic inference passes. The resulting variability is negligible overall, and is only marginally visible for the VGG11 models. Consequently, our observations are not affected by this.

Figure~\ref{fig:pruning-comparison} additionally shows the recorded MACE values over the percentage of retained weights. For ResNet and ViT models MACE decreases at high sparsities, which reflects an increased confidence about the models lowering accuracy. For VGG models, MACE is constant over the percentage of remaining weights. Overall, MACE trends are largely independant of pruning criterion, with the exception of the square pruning function in ViT models, where the large decrease in MACE at higher sparsity coincides with a drop off in accuracy.

The LTs generated through the IMP process have satisfactory prediction accuracy up to 90\%, achieving equal performance compared to the dense model in almost all cases. Given that, we try to seek a deeper understanding of how a winning LT is structured, such that it achieves good predictive performance with fewer parameters.

\section{What Defines a Lottery Ticket?}
\label{sec:sparsity-ratio}

\begin{figure*}[!t]
    \centering
    \includegraphics[width=0.33\linewidth]{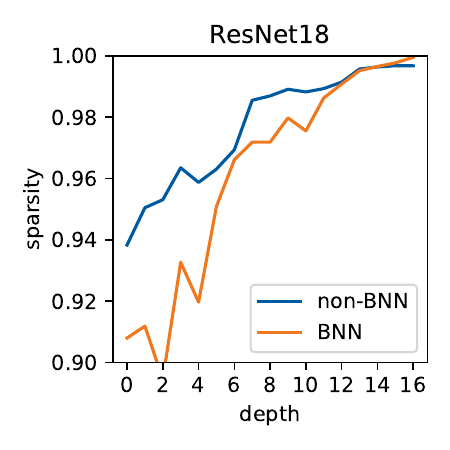}%
    \includegraphics[width=0.33\linewidth]{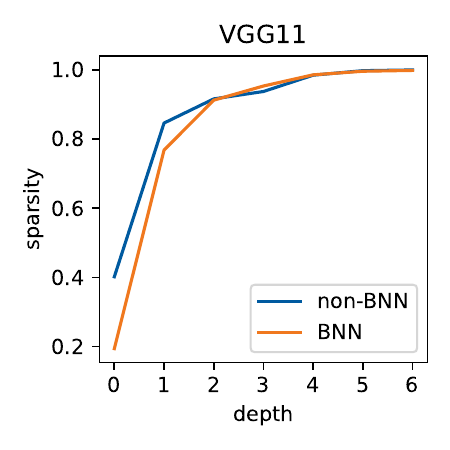}%
    \includegraphics[width=0.33\linewidth]{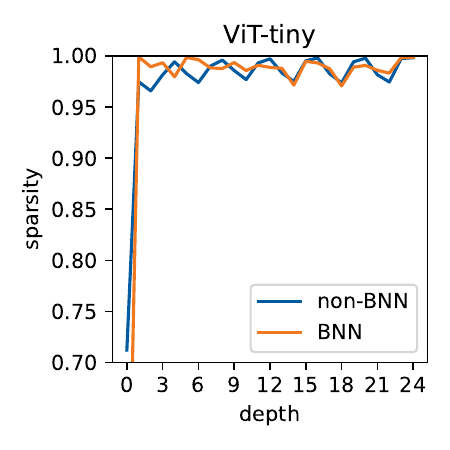}
    \caption{Layer-wise sparsity vs. layer index for ResNet (left), VGG (center), and ViT (right) training and pruning 20 times. The x-axis denotes the layer index, and the y-axis shows the resulting sparsity in each layer. Each plot shows the non-Bayesian model and the Bayesian model.}
    \label{fig:sparsity-ratios}
\end{figure*}

To answer the question of how a winning ticket is different from any other pruned network of the same sparsity, we ask ourselves the question \textit{which weights are pruned during the train-prune-reset cycle?} We investigate this by focusing on the last level, i.e., the models with the highest sparsity pruned with the \( s_{SNR} \) scoring function. Due to the unstructured nature of global pruning, node-level analysis is not useful. Instead, we compute the layer-wise sparsity ratio to see which layers are pruned to what amount. Figure~\ref{fig:sparsity-ratios} shows the sparsity ratios for all three trained models in their Bayesian and non-Bayesian variations. We observe that deeper layers get pruned more; these layers commonly have more parameters as compared to shallower layers, and thus, offer more potential to prune parameters. Comparing the Bayesian models to the non-Bayesian counterpart, this skew is even more pronounced. Parameters in the earlier layers are kept in the high sparsity regime. Notably, one can see a modulation in the curve for the VisionTransformer model. This corresponds to the architecture of stacked attention and MLP layers.

We have shown that sparse models could replace the dense, parameter-heavy models. However, as with all sparsity, the question arises whether these winning LTs can only be obtained through the train-prune-reset cycle, or also through a less computationally intensive process, e.g. through randomly pruning a network even before training. It is therefore of interest to study how a lottery ticket is different from any random ticket of the same sparsity, and how such a random ticket would perform.

\section{Reinitialization and Shuffling}
\begin{figure}
    \centering
    \includegraphics[width=0.95\linewidth]{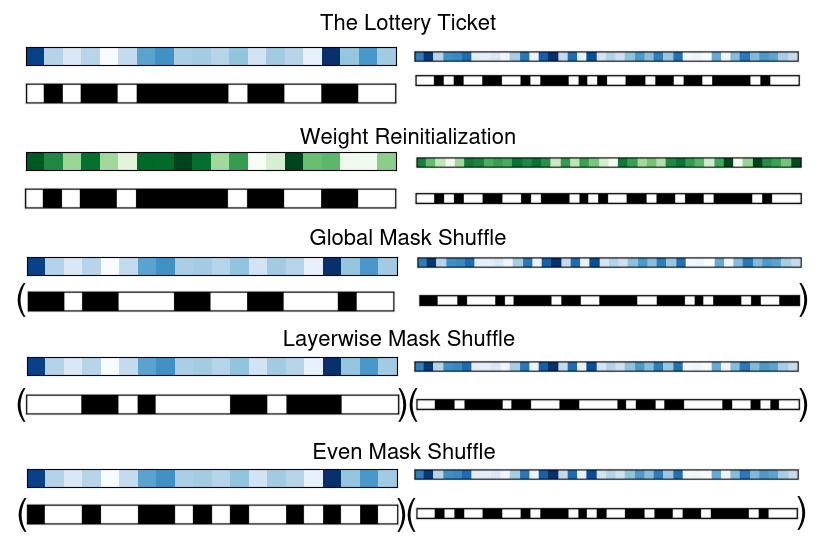}
    \caption{Visualization of the different randomizations applied to the weights (magnitude symbolized through color) and masks (binary, i.e. black and white) of a winning ticket.}
    \label{fig:lottery-viz}
\end{figure}
We use the winning tickets we found in ~\Cref{sec:pruning-bnn} to construct several other ``random'' weight-mask initializations, see Figure~\ref{fig:lottery-viz}. For a given weight-mask pair of any level, we apply \textit{weight reinitialization} or \textit{mask shuffling} and train as described in the training curriculum. For weight reinitialization, the mask is kept fixed, and we redraw all weights from the initial Kaiming normal distribution. For mask shuffling, the weight vector is fixed, and we test three different shuffling strategies for the mask and compare them. In \textbf{global shuffling}, all mask vectors of all layers are flattened and concatenated. Then a random permutation is applied, and the mask is restored to its original shape. This can lead to very uneven sparsity distributions across the layers. Due to their size difference, a small layer might be completely pruned.
For \textbf{even shuffling}, each layer is assigned a random binary mask with sparsity equal to the mean global sparsity. This alleviates the problem of pruned layers, but neglects that with global pruning, each layer has its own sparsity ratio as seen in~\Cref{sec:sparsity-ratio}.
Therefore, \textbf{layer-wise shuffling} redraws the mask for each layer according to its sparsity in the winning ticket found with IMP. Note that this information is only available due to the iterative process.

\begin{figure*}[!t]
    \centering
    \includegraphics[width=0.3\linewidth]{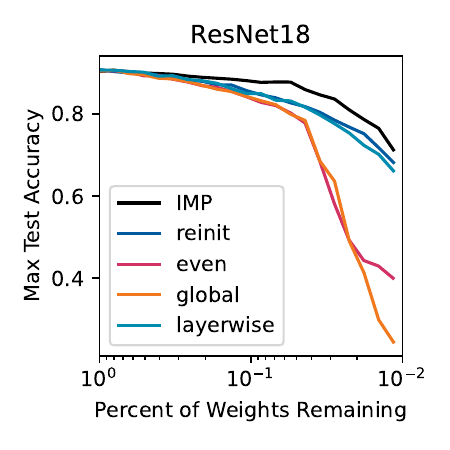}%
    \includegraphics[width=0.3\linewidth]{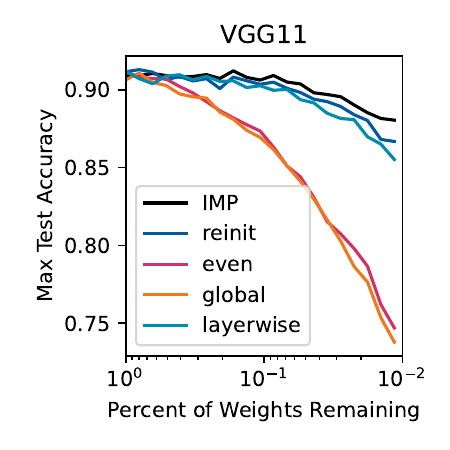}%
    \includegraphics[width=0.3\linewidth]{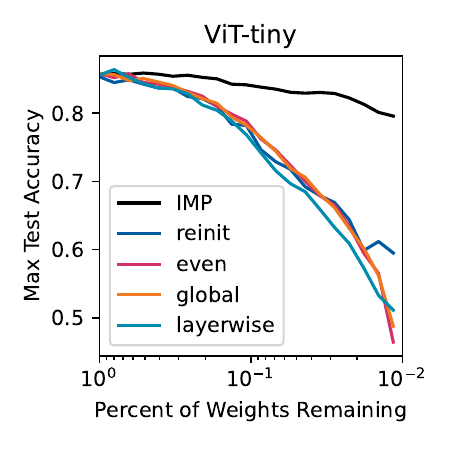}
    \includegraphics[width=0.3\linewidth]{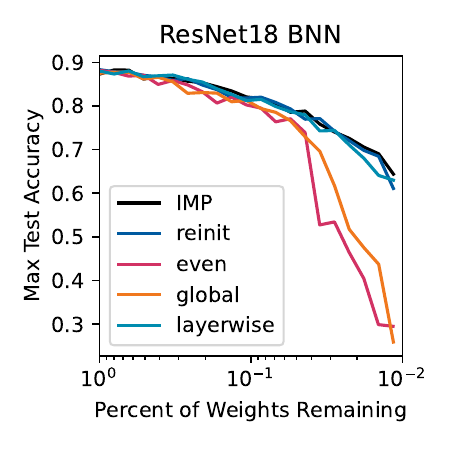}%
    \includegraphics[width=0.3\linewidth]{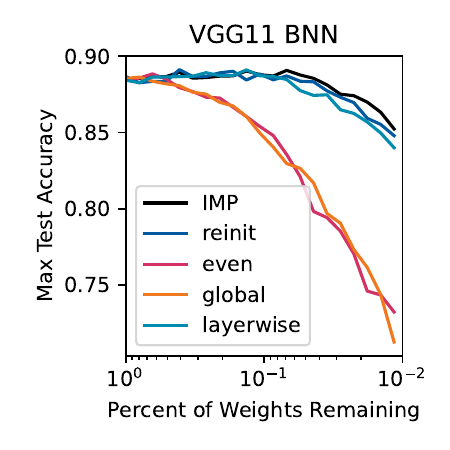}%
    \includegraphics[width=0.3\linewidth]{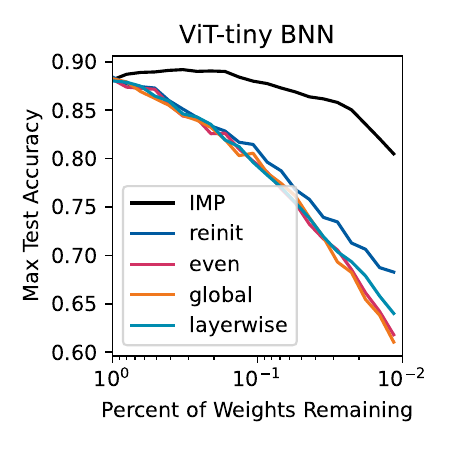}
    \caption{Test accuracy vs. percentage of weights remaining (log scale) for ResNet (left), VGG (middle), and ViT (right) trained on CIFAR10. The black line denotes the original lottery ticket obtained through IMP, the colored lines represent the accuracy achieved by reinitializing or shuffling and training the ticket of that sparsity level. The Bayesian and non-Bayesian models are shown in separate plots.}
    \label{fig:reshuffle}
\end{figure*}

Figure~\ref{fig:reshuffle} shows the peak test accuracy over the percentage of remaining weights for all reinitialization and shuffling strategies. Additionally, we show the result of the same reinitialization and shuffling done with non-Bayesian models, replicating the original LTH experiment.

The IMP curves are the ceiling in all configurations, demonstrating that the winning tickets found with IMP are superior to all others. The winning LTs outperforming random tickets confirms that both weight initialization and mask structure make up the sparse subnetwork identified by the LTH. For the ViT-tiny model, this gap between random and winning ticket is the largest, while the accuracy of all other network initializations drops off synchronously with increasing sparsity. In all cases, even and global shuffling removes any advantage a ticket retains over the train-prune-reset cycle. For ResNet18 and VGG11, layerwise shuffling and reinitializing the weights leads to comparable accuracy to the winning ticket. Because reinitialization of weights also keeps layerwise sparsity ratios equal, \emph{the layerwise ratios seem to be a crucial ingredient to a winning ticket}. On a smaller scale, weight reinitialization outperforms layerwise shuffling; therefore, the more fine-grained graph structure of a network might be the deciding factor over the coarse layerwise sparsity ratio.
For the ViT models, \emph{only the combination of weight initialization and mask produces highly performing networks}. This effect has to be further investigated, especially taking into account that large ViT models are commonly pre-trained on large image datasets like ImageNet~\citep{deng_imagenet_2009}. We hypothesize that this is an effect of the architectural design of attention-based transformers, where no inductive bias towards images is built into like in convolutional neural networks. 
We provide additional ablation experiments to model size, learning rate schedule and learning rate resetting in the appendix.

\section{Lottery Ticket Transplantation}
\label{sec:transplant}

\begin{figure*}[!t]
    \centering
    \includegraphics[width=0.3\linewidth]{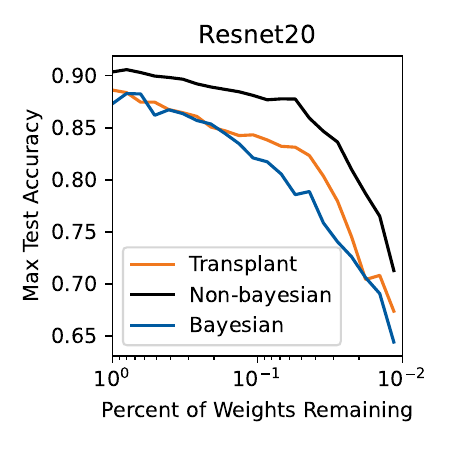}%
    \includegraphics[width=0.3\linewidth]{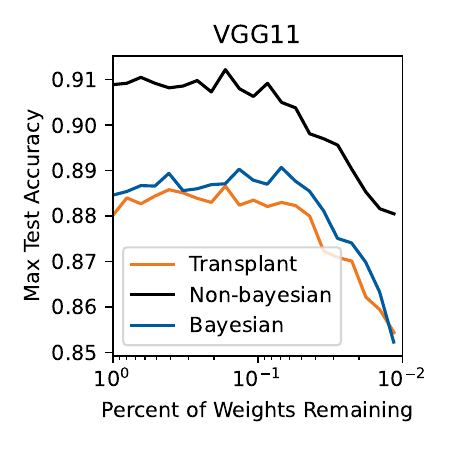}%
    \includegraphics[width=0.3\linewidth]{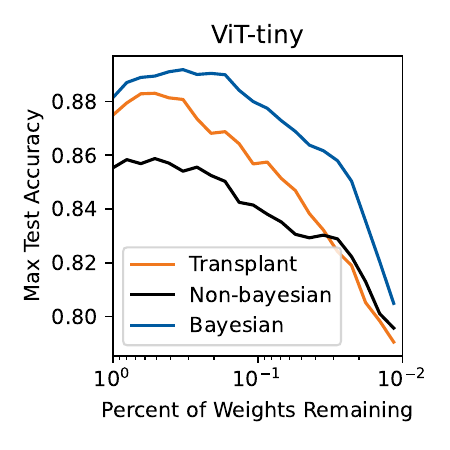}
    \caption{Test accuracy vs. percentage of weights remaining (log scale) for ResNet (left), VGG (middle), and ViT (right) trained on CIFAR10. Shown are LTs generated through deterministic IMP, a fully Bayesian approach, and the transplantation method.
    }
    \label{fig:transplant}
\end{figure*}

We use the generated LTs of the non-Bayesian deterministic models produced in Section~\ref{sec:pruning-bnn}, to initialize BNNs at matching sparsity, with the goal of obtaining performant Bayesian LTs while avoiding the full computational cost of Bayesian LT discovery. For each LT, we construct a BNN by copying the pruned weights into the posterior mean parameters $\mu$ and reusing the sparsity mask, while keeping the $\sigma$ parameters at their initialization. The transplanted tickets are then trained with a final VI phase instead of deterministic training. The resulting maximum test set performance is shown in Figure~\ref{fig:transplant}. For comparison, the performances of fully Bayesian LTs and non-bayesian LTs are shown.
Empirically, transplantation yields accuracies comparable to deterministic and Bayesian LTs for ResNet and VGG, but falls short the non-Bayesian performance in the case of ViT-tiny.

The approach reduces computational demand substantially, as VI training is approximately 3-$7\times$ more expensive than for deterministic models. Appendix results further show that transplanted models retain improved predictive calibration.

Table~\ref{tab:runtime} reports averaged observed training run times in minutes for both non-Bayesian and Bayesian models. The Bayesian models of VGG and ViT take roughly 5 times longer. Because the IMP process chains up to 20 trainings of a model together with a pruning stage, the proposed transplantation saves up to 50\% of the compute time.
\begin{table}[H]
    \centering
    \caption{Average runtime for a singular training level of models trained in Section~\ref{sec:pruning-bnn} in minutes.}
    \begin{tabularx}{\linewidth}{l c c c}
        \toprule
         & ResNet20 & VGG11 & ViT-Tiny\\
         \midrule
         Deterministic & 12 & 13 & 9 \\
        Bayesian & 33 & 72 & 51 \\
          \bottomrule
    \end{tabularx}
    \label{tab:runtime}
\end{table}

\section{Discussion}
\label{sec:discussion}
BNNs bring uncertainty quantification, robustness, and increased generalization capabilities to deep learning, but suffer from an increased compute demand, which could be alleviated through sparsity.
The Lottery Ticket Hypothesis presents a vital step towards empirical evidence and justification that training well-performing sparse neural networks is possible, and is therefore of interest to the field of BNNs. 
The LTH was shown to be effective in all kinds of models and tasks, but so far only for deterministic non-BNNs. We therefore explored the LTH in BNNs trained with variational inference, to evaluate the existance of sparse subnetworks as means to alleviate computational demand. We applied IMP to Bayesian models of ResNet, VGG and ViT by means of various pruning functions, and analysed the resulting winning tickets in terms of their layerwise sparsity ratio as well as performance compared to randomly redrawn and shuffled tickets.
For consistency with prior work, our focus is the CIFAR10 dataset, presenting an ideal testbed for understanding pruning dynamics in different architectures.

Similar to standard neural networks, we observe so-called winning tickets, which either match or outperform their dense counterpart. 
Notably, there are differences per model, which are likely related to the architecture. ResNet and VGG are convolution-based models, while the ViT relies on attention. The convolutional layer carries an inductive bias towards image input, while the transformer has to learn such a bias. Both ResNet and VGG winning tickets gain most of their performance through the right layerwise sparsity ratio and mask structure. But even so, the winning ticket found through IMP outperforms any reshuffled ticket and remains optimal. As VGG seems to be slightly more stable for training when pruned, we hypothesize that the residual connections of ResNet interfere when only some of the parameters in a kernel are pruned. The ViT model is additionally highly sensitive to the initial weights chosen, and hence, clear winning tickets can be observed, mirroring deterministic Transformers, where similar performance changes depending on initialization have been observed. We therefore stress our usage of the VisionTransformer model, which is usually pre-trained on larger datasets. 

We find that the iterative pruning scheme is largely independent of the used weight scoring function. Only in high sparsity regimes and in the ViT we observe differences. The posterior mean seems to dominate the pruning decision, and can already deliver good ticket performance. 

By plotting the sparsity ratio per layer of highly sparse winning tickets, we observe that deeper layers are pruned more. Compared to standard NNs, this effect is increased, possibly due to a higher uncertainty of weights that increases with depth in the VI framework. This could also be a reason for the increased robustness of BNNs.

We show that it is possible to transplant Lottery Tickets found in non-Bayesian NNs to almost matching accuracy in BNNs. For high sparsities this cuts training time more than in half, while retaining better than randomly pruned performance with an equally well calibration. This demonstrates that a full Bayesian training might not be needed in limited compute environments.

Lastly, we outline the limitations of our study. Due to the large compute demand of BNNs, we have limited our experiments to the dataset used in prior work, CIFAR10, a mid-sized image dataset. We leave the exploration of the LTH in BNNs trained on larger datasets for future work. Similarly, we used VI to implement our BNNs. While this is a common choice in the literature, there are other ways to quantify the uncertainty of a neural network. It is of interest whether our findings can be validated using other methods like MCMC for variational gradients or MC Dropout. Similarly to standard NNs, it is of interest whether completely sparse training with structured sparsity can be realized~\citep{evci_rigging_2021,lasby_dynamic_2024}.

\bibliography{sources}

@misc{lotteryticket,
      title={The Lottery Ticket Hypothesis: Finding Sparse, Trainable Neural Networks}, 
      author={Jonathan Frankle and Michael Carbin},
      year={2019},
      eprint={1803.03635},
      archivePrefix={arXiv},
      primaryClass={cs.LG},
      url={https://arxiv.org/abs/1803.03635}, 
}

@misc{aidriving,
      title={Recent Advancements in End-to-End Autonomous Driving using Deep Learning: A Survey}, 
      author={Pranav Singh Chib and Pravendra Singh},
      year={2023},
      eprint={2307.04370},
      archivePrefix={arXiv},
      primaryClass={cs.RO},
      url={https://arxiv.org/abs/2307.04370}, 
}

@misc{blundell2015,
      title={Weight Uncertainty in Neural Networks}, 
      author={Charles Blundell and Julien Cornebise and Koray Kavukcuoglu and Daan Wierstra},
      year={2015},
      eprint={1505.05424},
      archivePrefix={arXiv},
      primaryClass={stat.ML},
      url={https://arxiv.org/abs/1505.05424}, 
}

@inproceedings{practicalVariational,
 author = {Graves, Alex},
 booktitle = {Advances in Neural Information Processing Systems},
 editor = {J. Shawe-Taylor and R. Zemel and P. Bartlett and F. Pereira and K.Q. Weinberger},
 pages = {},
 publisher = {Curran Associates, Inc.},
 title = {Practical Variational Inference for Neural Networks},
 url = {https://proceedings.neurips.cc/paper_files/paper/2011/file/7eb3c8be3d411e8ebfab08eba5f49632-Paper.pdf},
 volume = {24},
 year = {2011}
}

@misc{stochasticVariational,
      title={Stochastic Variational Inference}, 
      author={Matt Hoffman and David M. Blei and Chong Wang and John Paisley},
      year={2013},
      eprint={1206.7051},
      archivePrefix={arXiv},
      primaryClass={stat.ML},
      url={https://arxiv.org/abs/1206.7051}, 
}

@article{paszke2018pytorch,
  author       = {Adam Paszke and
                  Sam Gross and
                  Francisco Massa and
                  Adam Lerer and
                  James Bradbury and
                  Gregory Chanan and
                  Trevor Killeen and
                  Zeming Lin and
                  Natalia Gimelshein and
                  Luca Antiga and
                  Alban Desmaison and
                  Andreas K{\"{o}}pf and
                  Edward Z. Yang and
                  Zach DeVito and
                  Martin Raison and
                  Alykhan Tejani and
                  Sasank Chilamkurthy and
                  Benoit Steiner and
                  Lu Fang and
                  Junjie Bai and
                  Soumith Chintala},
  title        = {PyTorch: An Imperative Style, High-Performance Deep Learning Library},
  journal      = {CoRR},
  volume       = {abs/1912.01703},
  year         = {2019},
  url          = {http://arxiv.org/abs/1912.01703},
  eprinttype    = {arXiv},
  eprint       = {1912.01703},
  bibsource    = {dblp computer science bibliography, https://dblp.org}
}

@misc{he2015deepresiduallearningimage,
      title={Deep Residual Learning for Image Recognition}, 
      author={Kaiming He and Xiangyu Zhang and Shaoqing Ren and Jian Sun},
      year={2015},
      eprint={1512.03385},
      archivePrefix={arXiv},
      primaryClass={cs.CV},
      url={https://arxiv.org/abs/1512.03385}, 
}

@misc{simonyan2015deepconvolutionalnetworkslargescale,
      title={Very Deep Convolutional Networks for Large-Scale Image Recognition}, 
      author={Karen Simonyan and Andrew Zisserman},
      year={2015},
      eprint={1409.1556},
      archivePrefix={arXiv},
      primaryClass={cs.CV},
      url={https://arxiv.org/abs/1409.1556}, 
}

@misc{saul1996meanfieldtheorysigmoid,
      title={Mean Field Theory for Sigmoid Belief Networks}, 
      author={L. K. Saul and T. Jaakkola and M. I. Jordan},
      year={1996},
      eprint={cs/9603102},
      archivePrefix={arXiv},
      primaryClass={cs.AI},
      url={https://arxiv.org/abs/cs/9603102}, 
}

@misc{wenzel_how_2020,
    title = {How {Good} is the {Bayes} {Posterior} in {Deep} {Neural} {Networks} {Really}?},
    url = {http://arxiv.org/abs/2002.02405},
    doi = {10.48550/arXiv.2002.02405},
    urldate = {2025-07-21},
    publisher = {arXiv},
    author = {Wenzel, Florian and Roth, Kevin and Veeling, Bastiaan S. and Świątkowski, Jakub and Tran, Linh and Mandt, Stephan and Snoek, Jasper and Salimans, Tim and Jenatton, Rodolphe and Nowozin, Sebastian},
    month = jul,
    year = {2020},
    note = {arXiv:2002.02405 [stat]},
}

@misc{hoefler_sparsity_2021,
    title = {Sparsity in {Deep} {Learning}: {Pruning} and growth for efficient inference and training in neural networks},
    shorttitle = {Sparsity in {Deep} {Learning}},
    url = {http://arxiv.org/abs/2102.00554},
    doi = {10.48550/arXiv.2102.00554},
    urldate = {2023-04-24},
    publisher = {arXiv},
    author = {Hoefler, Torsten and Alistarh, Dan and Ben-Nun, Tal and Dryden, Nikoli and Peste, Alexandra},
    month = jan,
    year = {2021},
    note = {arXiv:2102.00554 [cs]},
}

@misc{liu_survey_2024,
    title = {A {Survey} of {Lottery} {Ticket} {Hypothesis}},
    url = {http://arxiv.org/abs/2403.04861},
    doi = {10.48550/arXiv.2403.04861},
    urldate = {2024-04-25},
    publisher = {arXiv},
    author = {Liu, Bohan and Zhang, Zijie and He, Peixiong and Wang, Zhensen and Xiao, Yang and Ye, Ruimeng and Zhou, Yang and Ku, Wei-Shinn and Hui, Bo},
    month = mar,
    year = {2024},
    note = {arXiv:2403.04861 [cs]},
}

@misc{yu_playing_2020,
    title = {Playing the lottery with rewards and multiple languages: lottery tickets in {RL} and {NLP}},
    shorttitle = {Playing the lottery with rewards and multiple languages},
    url = {http://arxiv.org/abs/1906.02768},
    doi = {10.48550/arXiv.1906.02768},
    urldate = {2025-07-31},
    publisher = {arXiv},
    author = {Yu, Haonan and Edunov, Sergey and Tian, Yuandong and Morcos, Ari S.},
    month = feb,
    year = {2020},
    note = {arXiv:1906.02768 [stat]},
}

@misc{chen_earlybert_2021,
    title = {{EarlyBERT}: {Efficient} {BERT} {Training} via {Early}-bird {Lottery} {Tickets}},
    shorttitle = {{EarlyBERT}},
    url = {http://arxiv.org/abs/2101.00063},
    doi = {10.48550/arXiv.2101.00063},
    urldate = {2025-04-15},
    publisher = {arXiv},
    author = {Chen, Xiaohan and Cheng, Yu and Wang, Shuohang and Gan, Zhe and Wang, Zhangyang and Liu, Jingjing},
    month = jun,
    year = {2021},
    note = {arXiv:2101.00063 [cs]},
}

@misc{chen_unified_2021,
    title = {A {Unified} {Lottery} {Ticket} {Hypothesis} for {Graph} {Neural} {Networks}},
    url = {http://arxiv.org/abs/2102.06790},
    doi = {10.48550/arXiv.2102.06790},
    urldate = {2024-02-08},
    publisher = {arXiv},
    author = {Chen, Tianlong and Sui, Yongduo and Chen, Xuxi and Zhang, Aston and Wang, Zhangyang},
    month = jun,
    year = {2021},
    note = {arXiv:2102.06790 [cs, stat]},
}

@inproceedings{itahara_lottery_2020,
    title = {Lottery {Hypothesis} based {Unsupervised} {Pre}-training for {Model} {Compression} in {Federated} {Learning}},
    url = {http://arxiv.org/abs/2004.09817},
    doi = {10.1109/VTC2020-Fall49728.2020.9348439},
    urldate = {2025-07-31},
    booktitle = {2020 {IEEE} 92nd {Vehicular} {Technology} {Conference} ({VTC2020}-{Fall})},
    author = {Itahara, Sohei and Nishio, Takayuki and Morikura, Masahiro and Yamamoto, Koji},
    month = nov,
    year = {2020},
    note = {arXiv:2004.09817 [cs]},
    pages = {1--5},
}

@misc{renda_comparing_2020,
    title = {Comparing {Rewinding} and {Fine}-tuning in {Neural} {Network} {Pruning}},
    url = {http://arxiv.org/abs/2003.02389},
    doi = {10.48550/arXiv.2003.02389},
    urldate = {2024-09-13},
    publisher = {arXiv},
    author = {Renda, Alex and Frankle, Jonathan and Carbin, Michael},
    month = mar,
    year = {2020},
    note = {arXiv:2003.02389 [cs, stat]},
}

@misc{evci_rigging_2021,
    title = {Rigging the {Lottery}: {Making} {All} {Tickets} {Winners}},
    shorttitle = {Rigging the {Lottery}},
    url = {http://arxiv.org/abs/1911.11134},
    doi = {10.48550/arXiv.1911.11134},
    urldate = {2025-04-09},
    publisher = {arXiv},
    author = {Evci, Utku and Gale, Trevor and Menick, Jacob and Castro, Pablo Samuel and Elsen, Erich},
    month = jul,
    year = {2021},
    note = {arXiv:1911.11134 [cs]},
}

@misc{dosovitskiy_image_2021,
    title = {An {Image} is {Worth} 16x16 {Words}: {Transformers} for {Image} {Recognition} at {Scale}},
    shorttitle = {An {Image} is {Worth} 16x16 {Words}},
    url = {http://arxiv.org/abs/2010.11929},
    doi = {10.48550/arXiv.2010.11929},
    urldate = {2023-06-12},
    publisher = {arXiv},
    author = {Dosovitskiy, Alexey and Beyer, Lucas and Kolesnikov, Alexander and Weissenborn, Dirk and Zhai, Xiaohua and Unterthiner, Thomas and Dehghani, Mostafa and Minderer, Matthias and Heigold, Georg and Gelly, Sylvain and Uszkoreit, Jakob and Houlsby, Neil},
    month = jun,
    year = {2021},
    note = {arXiv:2010.11929 [cs]},
}

@article{krizhevsky2009learning,
  title={Learning multiple layers of features from tiny images},
  author={Krizhevsky, Alex and Hinton, Geoffrey and others},
  year={2009},
  publisher={Toronto, ON, Canada},
  journal={}
}

@inproceedings{izmailov_what_2021,
    title = {What {Are} {Bayesian} {Neural} {Network} {Posteriors} {Really} {Like}?},
    url = {https://proceedings.mlr.press/v139/izmailov21a.html},
    language = {en},
    urldate = {2025-08-25},
    booktitle = {Proceedings of the 38th {International} {Conference} on {Machine} {Learning}},
    publisher = {PMLR},
    author = {Izmailov, Pavel and Vikram, Sharad and Hoffman, Matthew D. and Wilson, Andrew Gordon Gordon},
    month = jul,
    year = {2021},
    note = {ISSN: 2640-3498},
    pages = {4629--4640},
}

@misc{nemeth_stochastic_2019,
    title = {Stochastic gradient {Markov} chain {Monte} {Carlo}},
    url = {http://arxiv.org/abs/1907.06986},
    doi = {10.48550/arXiv.1907.06986},
    urldate = {2025-08-26},
    publisher = {arXiv},
    author = {Nemeth, Christopher and Fearnhead, Paul},
    month = jul,
    year = {2019},
    note = {arXiv:1907.06986 [stat]},
}

@article{jantre_spike-and-slab_2025,
    title = {Spike-and-{Slab} {Shrinkage} {Priors} for {Structurally} {Sparse} {Bayesian} {Neural} {Networks}},
    volume = {36},
    issn = {2162-2388},
    url = {https://ieeexplore.ieee.org/abstract/document/10740530},
    doi = {10.1109/TNNLS.2024.3485529},
    number = {6},
    urldate = {2025-08-26},
    journal = {IEEE Transactions on Neural Networks and Learning Systems},
    author = {Jantre, Sanket and Bhattacharya, Shrijita and Maiti, Tapabrata},
    month = jun,
    year = {2025},
    pages = {11176--11188},
}

@misc{ghosh_model_2017,
    title = {Model {Selection} in {Bayesian} {Neural} {Networks} via {Horseshoe} {Priors}},
    url = {http://arxiv.org/abs/1705.10388},
    doi = {10.48550/arXiv.1705.10388},
    urldate = {2025-08-26},
    publisher = {arXiv},
    author = {Ghosh, Soumya and Doshi-Velez, Finale},
    month = may,
    year = {2017},
    note = {arXiv:1705.10388 [stat]},
}

@inproceedings{carvalho_handling_2009,
    title = {Handling {Sparsity} via the {Horseshoe}},
    url = {https://proceedings.mlr.press/v5/carvalho09a.html},
    language = {en},
    urldate = {2025-08-26},
    booktitle = {Proceedings of the {Twelfth} {International} {Conference} on {Artificial} {Intelligence} and {Statistics}},
    publisher = {PMLR},
    author = {Carvalho, Carlos M. and Polson, Nicholas G. and Scott, James G.},
    month = apr,
    year = {2009},
    note = {ISSN: 1938-7228},
    pages = {73--80},
}

@misc{cherief-abdellatif_convergence_2019,
    title = {Convergence {Rates} of {Variational} {Inference} in {Sparse} {Deep} {Learning}},
    url = {http://arxiv.org/abs/1908.04847},
    doi = {10.48550/arXiv.1908.04847},
    urldate = {2025-08-26},
    publisher = {arXiv},
    author = {Chérief-Abdellatif, Badr-Eddine},
    month = sep,
    year = {2019},
    note = {arXiv:1908.04847 [math]},
}

@misc{mathew_pruning_2023,
    title = {Pruning a neural network using {Bayesian} inference},
    url = {http://arxiv.org/abs/2308.02451},
    doi = {10.48550/arXiv.2308.02451},
    urldate = {2025-08-26},
    publisher = {arXiv},
    author = {Mathew, Sunil and Rowe, Daniel B.},
    month = aug,
    year = {2023},
    note = {arXiv:2308.02451 [stat]},
}

@misc{ko_magnitude_2019,
    title = {Magnitude and {Uncertainty} {Pruning} {Criterion} for {Neural} {Networks}},
    url = {http://arxiv.org/abs/1912.04845},
    doi = {10.48550/arXiv.1912.04845},
    urldate = {2025-08-26},
    publisher = {arXiv},
    author = {Ko, Vinnie and Oehmcke, Stefan and Gieseke, Fabian},
    month = dec,
    year = {2019},
    note = {arXiv:1912.04845 [cs]},
}

@misc{li_training_2024,
    title = {Training {Bayesian} {Neural} {Networks} with {Sparse} {Subspace} {Variational} {Inference}},
    url = {http://arxiv.org/abs/2402.11025},
    doi = {10.48550/arXiv.2402.11025},
    urldate = {2025-08-26},
    publisher = {arXiv},
    author = {Li, Junbo and Miao, Zichen and Qiu, Qiang and Zhang, Ruqi},
    month = feb,
    year = {2024},
    note = {arXiv:2402.11025 [cs]
version: 1},
}

@misc{wu_functional_2024,
    title = {Functional {Stochastic} {Gradient} {MCMC} for {Bayesian} {Neural} {Networks}},
    url = {http://arxiv.org/abs/2409.16632},
    doi = {10.48550/arXiv.2409.16632},
    urldate = {2025-08-26},
    publisher = {arXiv},
    author = {Wu, Mengjing and Xuan, Junyu and Lu, Jie},
    month = oct,
    year = {2024},
    note = {arXiv:2409.16632 [cs]},
}

@misc{kim_learning_2024,
    title = {Learning to {Explore} for {Stochastic} {Gradient} {MCMC}},
    url = {http://arxiv.org/abs/2408.09140},
    doi = {10.48550/arXiv.2408.09140},
    urldate = {2025-08-26},
    publisher = {arXiv},
    author = {Kim, SeungHyun and Jung, Seohyeon and Kim, Seonghyeon and Lee, Juho},
    month = aug,
    year = {2024},
    note = {arXiv:2408.09140 [cs]},
}

@misc{faller_optimal_2025,
    title = {Optimal {Subspace} {Inference} for the {Laplace} {Approximation} of {Bayesian} {Neural} {Networks}},
    url = {http://arxiv.org/abs/2502.02345},
    doi = {10.48550/arXiv.2502.02345},
    urldate = {2025-08-26},
    publisher = {arXiv},
    author = {Faller, Josua and Martin, Jörg},
    month = feb,
    year = {2025},
    note = {arXiv:2502.02345 [cs]},
}

@misc{kingma_variational_2015,
    title = {Variational {Dropout} and the {Local} {Reparameterization} {Trick}},
    url = {http://arxiv.org/abs/1506.02557},
    doi = {10.48550/arXiv.1506.02557},
    urldate = {2025-08-26},
    publisher = {arXiv},
    author = {Kingma, Diederik P. and Salimans, Tim and Welling, Max},
    month = dec,
    year = {2015},
    note = {arXiv:1506.02557 [stat]},
}

@misc{frankle_stabilizing_2020,
    title = {Stabilizing the {Lottery} {Ticket} {Hypothesis}},
    url = {http://arxiv.org/abs/1903.01611},
    doi = {10.48550/arXiv.1903.01611},
    urldate = {2025-08-29},
    publisher = {arXiv},
    author = {Frankle, Jonathan and Dziugaite, Gintare Karolina and Roy, Daniel M. and Carbin, Michael},
    month = jul,
    year = {2020},
    note = {arXiv:1903.01611 [cs]},
}

@misc{wang_picking_2020,
    title = {Picking {Winning} {Tickets} {Before} {Training} by {Preserving} {Gradient} {Flow}},
    url = {http://arxiv.org/abs/2002.07376},
    doi = {10.48550/arXiv.2002.07376},
    urldate = {2025-02-17},
    publisher = {arXiv},
    author = {Wang, Chaoqi and Zhang, Guodong and Grosse, Roger},
    month = aug,
    year = {2020},
    note = {arXiv:2002.07376 [cs]},
}

@misc{lee_snip_2019,
    title = {{SNIP}: {Single}-shot {Network} {Pruning} based on {Connection} {Sensitivity}},
    shorttitle = {{SNIP}},
    url = {http://arxiv.org/abs/1810.02340},
    doi = {10.48550/arXiv.1810.02340},
    urldate = {2025-04-14},
    publisher = {arXiv},
    author = {Lee, Namhoon and Ajanthan, Thalaiyasingam and Torr, Philip H. S.},
    month = feb,
    year = {2019},
    note = {arXiv:1810.02340 [cs]},
}

@misc{shen_sparse_2024,
    title = {Sparse {Deep} {Learning} {Models} with the \${\textbackslash}ell\_1\$ {Regularization}},
    url = {http://arxiv.org/abs/2408.02801},
    doi = {10.48550/arXiv.2408.02801},
    urldate = {2025-08-29},
    publisher = {arXiv},
    author = {Shen, Lixin and Wang, Rui and Xu, Yuesheng and Yan, Mingsong},
    month = aug,
    year = {2024},
    note = {arXiv:2408.02801 [cs]},
}

@misc{sui_elrt_2024,
    title = {{ELRT}: {Efficient} {Low}-{Rank} {Training} for {Compact} {Convolutional} {Neural} {Networks}},
    shorttitle = {{ELRT}},
    url = {http://arxiv.org/abs/2401.10341},
    doi = {10.48550/arXiv.2401.10341},
    urldate = {2025-04-10},
    publisher = {arXiv},
    author = {Sui, Yang and Yin, Miao and Gong, Yu and Xiao, Jinqi and Phan, Huy and Yuan, Bo},
    month = jan,
    year = {2024},
    note = {arXiv:2401.10341 [cs]},
}

@misc{elsen_fast_2019,
    title = {Fast {Sparse} {ConvNets}},
    url = {http://arxiv.org/abs/1911.09723},
    doi = {10.48550/arXiv.1911.09723},
    urldate = {2025-08-29},
    publisher = {arXiv},
    author = {Elsen, Erich and Dukhan, Marat and Gale, Trevor and Simonyan, Karen},
    month = nov,
    year = {2019},
    note = {arXiv:1911.09723 [cs]},
}

@misc{han_deep_2016,
    title = {Deep {Compression}: {Compressing} {Deep} {Neural} {Networks} with {Pruning}, {Trained} {Quantization} and {Huffman} {Coding}},
    shorttitle = {Deep {Compression}},
    url = {http://arxiv.org/abs/1510.00149},
    doi = {10.48550/arXiv.1510.00149},
    urldate = {2025-09-15},
    publisher = {arXiv},
    author = {Han, Song and Mao, Huizi and Dally, William J.},
    month = feb,
    year = {2016},
    note = {arXiv:1510.00149 [cs]},
}

@book{neal_mcmc_2011,
    title = {{MCMC} using {Hamiltonian} dynamics},
    url = {http://arxiv.org/abs/1206.1901},
    urldate = {2025-09-15},
    author = {Neal, Radford M.},
    month = may,
    year = {2011},
    doi = {10.1201/b10905},
    note = {arXiv:1206.1901 [stat]},
}

@inproceedings{deng_imagenet_2009,
    title = {{ImageNet}: {A} large-scale hierarchical image database},
    shorttitle = {{ImageNet}},
    url = {https://ieeexplore.ieee.org/document/5206848},
    doi = {10.1109/CVPR.2009.5206848},
    urldate = {2025-09-19},
    booktitle = {2009 {IEEE} {Conference} on {Computer} {Vision} and {Pattern} {Recognition}},
    author = {Deng, Jia and Dong, Wei and Socher, Richard and Li, Li-Jia and Li, Kai and Fei-Fei, Li},
    month = jun,
    year = {2009},
    note = {ISSN: 1063-6919},
    pages = {248--255},
}

@misc{lasby_dynamic_2024,
    title = {Dynamic {Sparse} {Training} with {Structured} {Sparsity}},
    url = {http://arxiv.org/abs/2305.02299},
    doi = {10.48550/arXiv.2305.02299},
    urldate = {2025-09-30},
    publisher = {arXiv},
    author = {Lasby, Mike and Golubeva, Anna and Evci, Utku and Nica, Mihai and Ioannou, Yani},
    month = feb,
    year = {2024},
    note = {arXiv:2305.02299 [cs]},
}

@inproceedings{noauthor_bayes_2025,
    title = {Bayes {Always} {Wins} the {Lottery} in {Monte} {Carlo}},
    url = {https://openreview.net/forum?id=Ilnbgf1eeS},
    language = {en},
    urldate = {2026-01-19},
    author = {Marsh, Peter and Ergun, Dario Deniz and Kuruoglu, Ercan Engin},
    month = oct,
    year = {2025},
}

@article{weyrauch_torch_blue_2026,
    title = {torch\_blue: {A} {Flexible} {Python} {Package} for {Bayesian} {Neural} {Networks} in {PyTorch}},
    volume = {11},
    issn = {2475-9066},
    shorttitle = {torch\_blue},
    url = {https://joss.theoj.org/papers/10.21105/joss.09415},
    doi = {10.21105/joss.09415},
    language = {en},
    number = {117},
    urldate = {2026-02-20},
    journal = {Journal of Open Source Software},
    author = {Weyrauch, Arvid and Heyen, Lars H. and Muriedas, Juan Pedro Gutiérrez Hermosillo and Hsia, Pei-Hsuan and Özdemir, Asena Karolin and Streit, Achim and Götz, Markus and Debus, Charlotte},
    month = jan,
    year = {2026},
    pages = {9415},
}

\appendix

\section{Ablation on Model Size and Learning Rate}
\begin{figure*}
    \centering
    \includegraphics[width=0.33\linewidth]{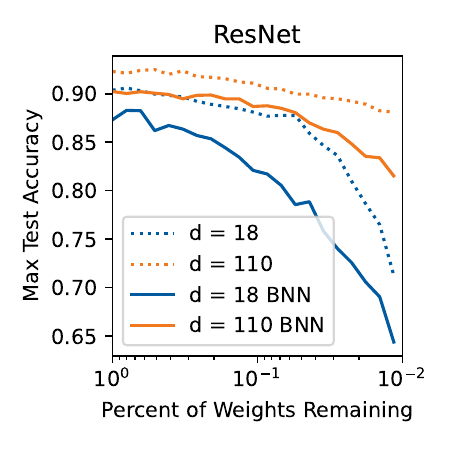}%
    \includegraphics[width=0.33\linewidth]{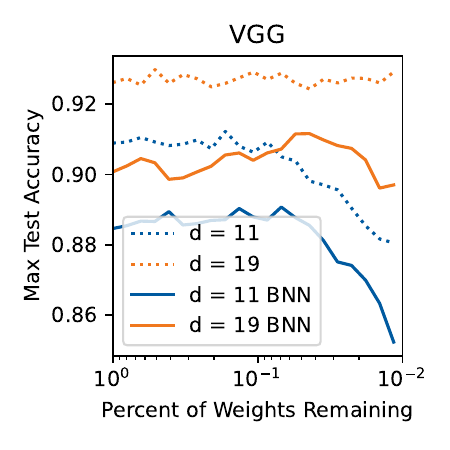}%
    \includegraphics[width=0.33\linewidth]{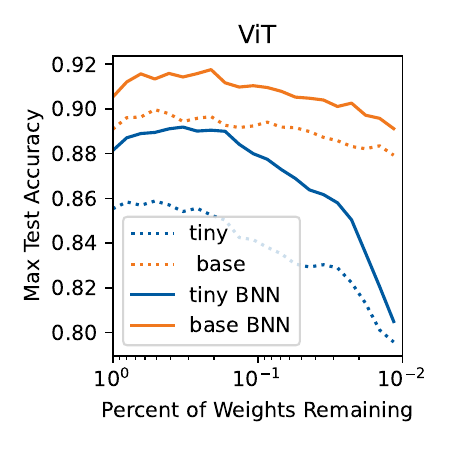}
    \caption{Test accuracy vs. Percentage of weights remaining (log scale) for ResNet, VGG, and ViT of different sizes trained on CIFAR10. Shown are non-Bayesian and Bayesian models.}
    \label{fig:model-size}
\end{figure*}

We compare the performance of differently sized models, scaling all models up. Both for non-Bayesian and Bayesian models, we perform the train-prune-reset cycle on the models ResNet110, VGG19, and ViT-base. We show the maximum test accuracy achieved over the percentage of remaining weights in Figure \ref{fig:model-size}. As expected, the models with a higher parameter count perform better. The qualitative behaviour with increasingly pruned parameters stays the same, with a drop in performance when reaching high sparsity. For the case of VGG19, both non-Bayesian and Bayesian, we do not see a drop in performance even at the highest sparsity ratio. This could be due to the heavy over-parametrization.

\cite{lotteryticket} found that winning tickets trained with a lower learning rate outperform tickets with a higher learning rate at high sparsity. Warmup helps to close the gap compared to the unpruned network. We therefore repeat these experiments for BNNs, comparing lottery tickets found with \( s_{SNR}\) pruning but with different base learning rates and additional warmup. Although we also find matching LTs with low learning rate and even surpassing LTs with warmup, this simply comes at a cost of less overall maximum accuracy. For medium sparsity, lottery tickets with warmup equal the high learning rate, but a high learning rate still produces the best performing tickets at high sparsity.

\begin{figure*}
    \centering
    \includegraphics[width=0.33\linewidth]{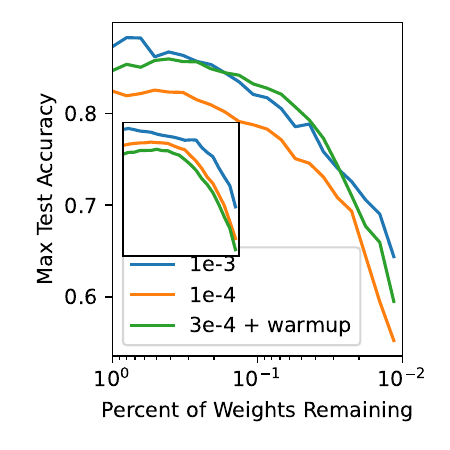}%
    \includegraphics[width=0.33\linewidth]{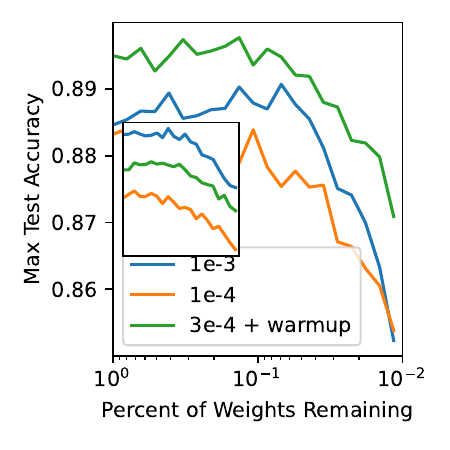}%
    \includegraphics[width=0.33\linewidth]{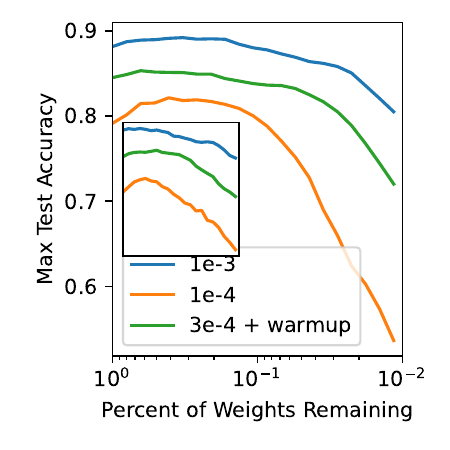}
    \caption{Test accuracy vs. Percentage of weights remaining (log scale) for ResNet, VGG, and ViT trained on CIFAR10. The colored lines represent different initial learning rates. Included as a subplot are the equivalent non-Bayesian models.}
    \label{fig:learning-rate}
\end{figure*}

\section{Does LRR outperform IMP?}

\begin{figure*}
    \centering
    \includegraphics[width=0.33\linewidth]{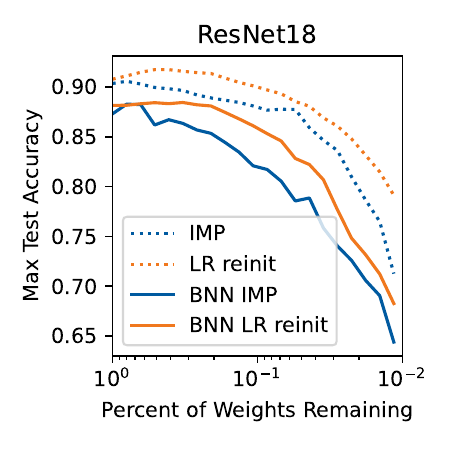}%
    \includegraphics[width=0.33\linewidth]{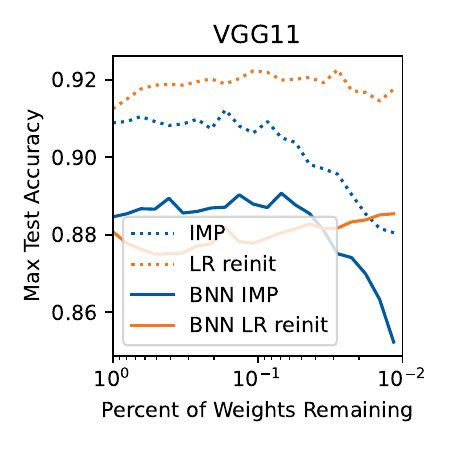}%
    \includegraphics[width=0.33\linewidth]{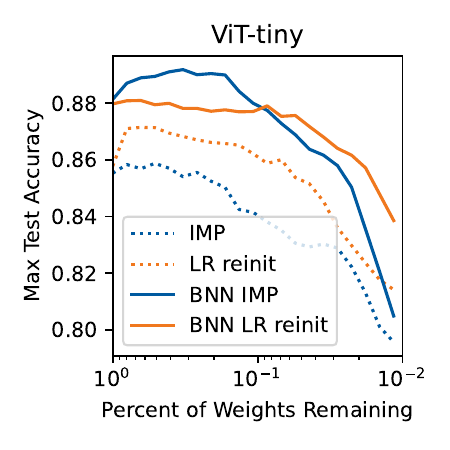}
    \caption{Test accuracy vs. percentage of weights remaining (log scale) for ResNet, VGG, and ViT trained on CIFAR10. The lines represent tickets obtained through either Learning Rate Rewinding or Iterative Magnitude Pruning. Shown are non-Bayesian and Bayesian models.}
    \label{fig:lr-comparison}
\end{figure*}

LRR~\citep{renda_comparing_2020} changes the way IMP is done. Instead of resetting weights still active after pruning to their initial value, training for the next sparsity level continues with the learned values. The learning rate schedule is reset instead. This was found to be a viable if not better alternative to IMP, especially in larger models. We briefly investigate how LRR performs in comparison to IMP in ResNet, VGG, and ViT, both in a Bayesian and non-Bayesian setting. We hypothesize LRR to outperform IMP in BNNs, because the exact pointwise initialization emphasized by IMP is far less meaningful in BNNs where parameters are distributions. Resetting to initial weights discards the learned posterior information that governs predictive uncertainty and makes the pruning mask sensitive to sampling noise. LRR focuses on the learning dynamics rather than a specific weight realization, making it better aligned with BNNs distributional nature.

Figure~\ref{fig:lr-comparison} shows the recorded maximum test accuracies achieved both with IMP and LRR.
For ResNet18 we see a clear improvement in LRR over IMP, across all sparsity ratios in the standard NN and the BNN model.
For VGG11, again LRR outperforms IMP across all sparsity levls, however, not in the Bayesian variant. Only at very high sparsity does the performance drop off for IMP while it does not for LRR, which has stable performance up to the highest sparsity level.
Pruning the bayesian variant of ViT-tiny, LRR outperforms IMP only at high sparsities. In the standard NN, LRR produces a higher accuracy at every sparsity level.
We summarize our findings: LRR achieves a higher predictive performance at high sparsity, and seems to produce more stable winning tickets. While it is sometimes outperformed by IMP at low sparsity, it often achieves matching or surpassing accuracy to the dense counterpart, even though it does not take the random starting weight initialization into account, as proposed by the LTH.

\section{Calibration Difference in Transplantation}
\begin{figure*}
    \centering
    \includegraphics[width=0.33\linewidth]{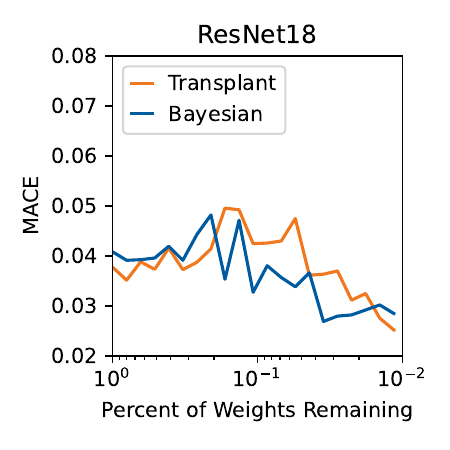}%
    \includegraphics[width=0.33\linewidth]{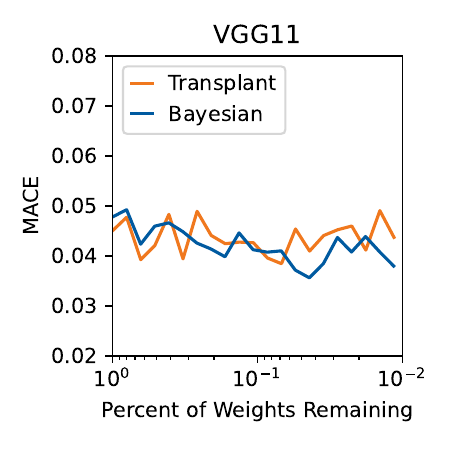}%
    \includegraphics[width=0.33\linewidth]{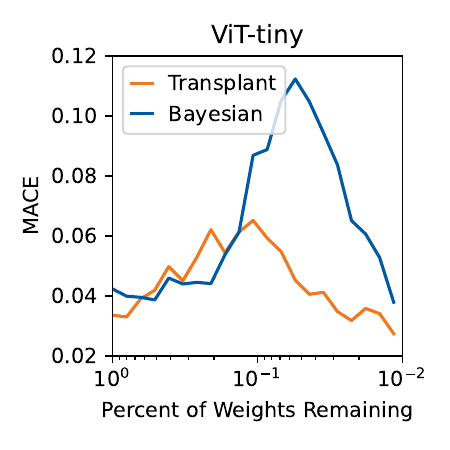}
    \caption{MACE vs. percentage of weights remaining for transplanted LTs (see Section~\ref{sec:transplant}) and fully Bayesian LTs.}
    \label{fig:mace_transplant}
\end{figure*}
We show the calibration error plots for models mentioned in Section~\ref{sec:transplant} in Figure~\ref{fig:mace_transplant}. We compare to the BNNs obtained through full Bayesian Training and IMP. MACE trends of transplanted tickets largely mirror the full Bayesian models. For ResNet and ViT this means a decrease in calibration error in the high sparsity regime, and a relatively constant MACE for VGG models.

\end{document}